\pdfoutput=1

\documentclass[11pt]{article}
\usepackage[table]{xcolor}
\usepackage[]{acl}

\usepackage{times}
\usepackage{latexsym}

\usepackage[T1]{fontenc}

\usepackage[utf8]{inputenc}

\usepackage{microtype}

\usepackage{todonotes}
\usepackage{url}
\usepackage{color}
\usepackage{makecell}
\usepackage{bbm}
\usepackage{amsmath}
\usepackage{arydshln}
\usepackage{subcaption}
\usepackage{caption}
\usepackage{multirow}
\usepackage{graphicx}
\usepackage[linesnumbered,ruled]{algorithm2e}

\definecolor{missing}{RGB}{255, 255, 0} 

\definecolor{best}{RGB}{255, 255, 0} 

\definecolor{muchworse}{RGB}{255, 125, 125} 
\newcommand{\muchworse}{\cellcolor{muchworse}}
\definecolor{muchbetter}{RGB}{125, 255, 125} 
\newcommand{\muchbetter}{\cellcolor{muchbetter}}

\definecolor{worse}{RGB}{255, 200, 200} 
\newcommand{\worse}{\cellcolor{worse}}
\definecolor{better}{RGB}{200, 255, 200} 
\newcommand{\better}{\cellcolor{better}}

\usepackage{tikz}
\def\checkmark{\tikz\fill[scale=0.4](0,.35) -- (.25,0) -- (1,.7) -- (.25,.15) -- cycle;}

\title{Por Qué Não Utiliser Alla Språk? \\Mixed Training with Gradient Optimization in \\Few-Shot Cross-Lingual Transfer}

\author{Haoran Xu, Kenton Murray\\[1em]
Johns Hopkins University\\
\texttt{\{hxu64,kenton\}@jhu.edu}\\[1em]
}

\begin{document}
\maketitle
\begin{abstract}
The current state-of-the-art for few-shot cross-lingual transfer learning first trains on abundant labeled data in the source language and then fine-tunes with a few examples on the target language, termed \textit{target-adapting}. Though this has been demonstrated to work on a variety of tasks, in this paper we show some deficiencies of this approach and propose a one-step mixed training method that trains on both source and target data with \textit{stochastic gradient surgery}, a novel gradient-level optimization. Unlike the previous studies that focus on one language at a time when target-adapting, we use one model to handle \textit{all target languages simultaneously} to avoid excessively language-specific models. Moreover, we discuss the unreality of utilizing large target development sets for model selection in previous literature. We further show that our method is both \textit{development-free} for target languages, and is also able to escape from overfitting issues. We conduct a large-scale experiment on 4 diverse NLP tasks across up to 48 languages. Our proposed method achieves state-of-the-art performance on all tasks and outperforms target-adapting by a large margin\footnote{Code is available at: \url{https://github.com/fe1ixxu/Mixed-Gradient-Few-Shot}.}, especially for languages that are linguistically distant from the source language, e.g., 7.36\% F1 absolute gain on average for the NER task, up to 17.60\% on Punjabi.
\end{abstract}
\section{Introduction}
The cost of linguistic data annotation and a plethora of differences across language resources and structures of natural language processing (NLP) tasks result in the problem that sufficient labeled data is only available for a handful of high-resource languages \citep{bender2011achieving}. The lack of data for low-resource languages leads to the need for effective cross-lingual transfer learning, which aims to leverage abundant labeled high-resource languages to improve model performance on low-resource ones. The majority of methods for cross-lingual transfer are mainly based on multilingual language models (LMs) \citep{devlin-etal-2019-bert,conneau-etal-2020-unsupervised,xue-etal-2021-mt5} which are pre-trained on massive multilingual data. \textit{Zero-shot cross-lingual transfer} is widely explored where a multilingual LM is trained on a large amount of labeled data in the source language without any target data, and then is directly evaluated on the target test set, frequently achieving surprisingly good performance \citep{wu-dredze-2019-beto,pires-etal-2019-multilingual,conneau-etal-2020-unsupervised}. Recently, \citet{lauscher-etal-2020-zero} emphasize the effective mechanism of \textit{few-shot cross-lingual transfer} for improving target-language performance, where only a few (such as 10) extra target examples can obtain substantial improvements. The current state-of-the-art methods for few-shot cross-lingual transfer learning \citep{lauscher-etal-2020-zero,hedderich-etal-2020-transfer,maurya-etal-2021-zmbart,zhao-etal-2021-closer} utilize the source-trained model (the same model training on the source data in zero-shot learning) to fine-tune on small target examples, which is termed \textit{target-adapting}.

In this paper, we dissect the potential weaknesses of the ubiquitous target-adapting method and propose a one-step mixed training method that trains on both source and target data with a novel gradient-level optimization, \textbf{stochastic gradient surgery}. Specifically, we highlight 6 benefits (contributions) of our method in this paper as follows:

\noindent \textbf{(1)} \textbf{State-of-The-Art Performance:}  Our proposed method achieves significant improvements compared to target-adapting on 4 diverse NLP tasks across up to 48 languages. For instance, averaged over all target languages, we demonstrate an absolute F1 improvement of 7.36\% on NER using 5-shot learning, with our best performance gains on Punjabi where the gap is 17.60\%  (Section \ref{sec:experiments}).

\noindent \textbf{(2)} \textbf{One Model for All Languages:}
The target-adapting step generally focuses on only one target language. With the proposed method, we do not need to fine-tune specialized models for every target language, which is of particular interest when scaling to dozens or even hundreds of languages. We discuss the benefits of mixed training one model on all target languages, even when their number of shots is extremely small (Section \ref{sec:all_target_lang}).

\noindent \textbf{(3)} \textbf{Efficient Gradient De-Conflicting and Information De-Dilution:}
Two issues arise when mixed training uses data from all target languages in addition to the source language --- conflicting gradients among languages and target information dilution. Stochastic gradient surgery efficiently de-conflicts gradients and de-dilutes the target information (Section \ref{sec:gradient-co-train} and \ref{sec:sgs}).

\noindent \textbf{(4) Single Language Friendly:} Though our proposed method normally uses information from multiple target languages, in the simplest setting, where we only have a single target language, stochastic gradient surgery trained on source and target still substantially outperforms standard target-adapting. The improvement is especially pronounced for languages linguistically distant from the source language (Section \ref{sec:cotrain-single}).

\noindent \textbf{(5) The Same Script Helps:} 
For a specific language, the model is able to use information learned from other languages. In Section \ref{sec:co-train-subset}, we show that this gain is most pronounced in languages that use the same script.

\noindent \textbf{(6) Development-Free for Target Languages:} 
Target development (dev) set used by previous studies \citep{hsu-etal-2019-zero,zhao-etal-2021-closer} significantly outnumber training examples in few-shot cross-lingual learning, which is not realistic in the true low-resource settings. However, target-adapting can be prone to overfit on small examples without target dev sets. In comparison, our proposed method is development-free for target languages and able to escape overfitting issues (Section \ref{sec:dev-free}).

\section{Background and Related Works}
\subsection{Cross-Lingual Transfer Learning}
Cross-lingual transfer learning enables systems to co-learn the meaning of words across languages and facilitates model transfer between languages, particularly from high-resource to low-resource languages \citep{ruder2019survey}, even for languages that are linguistically distant \citep{xu-etal-2021-gradual,yarmohammadi2021everything}. Language transfer is based on finding a shared cross-lingual space for source and target languages. One of the most common methods is to align the source and target embedding spaces, termed cross-lingual word embeddings (CLWEs) \citep{mikolov2013exploiting,artetxe-etal-2016-learning,lample2018word,vulic-etal-2019-really,xu2021cross}. Recently, multilingual pre-trained encoders have shown stronger effectiveness over CLWEs for cross-lingual transfer in various tasks \citep{10.1162/tacl_a_00288,wu-dredze-2019-beto}. While some studies utilize static pre-trained encoders for transfer learning \citep{wang-etal-2019-cross,xu-koehn-2021-zero}, the majority of studies continuously train encoders for cross-lingual transfer \citep{conneau-etal-2020-unsupervised,luo-etal-2021-veco,xue-etal-2021-mt5} based on the finding that source and target representations are still aligned after only fine-tuning on the source data \citep{hsu-etal-2019-zero}.

\subsection{Few-Shot Learning}
Few-shot learning was firstly investigated in computer vision \citep{fei2006one}. Currently, the majority of studies for NLP tasks are designed for one single language (usually English), e.g., model agnostic meta-learning \citep{pmlr-v70-finn17a} and prototypical networks \citep{snell2017prototypical}. However, limited few-shot studies are explored in cross-lingual settings. Recent works mainly focus on zero-shot cross-lingual transfer to evaluate the cross-lingual generalization capabilities of multilingual representations, e.g., XTREME \citep{hu2020xtreme,ruder-etal-2021-xtreme} and XGLUE \citep{liang-etal-2020-xglue}. \citet{lauscher-etal-2020-zero} further emphasize that additional fine-tuning on a few inexpensive labeled target-language instances is surprisingly effective across broad NLP tasks. \citet{zhao-etal-2021-closer} highlight the sensitivity of the selection of the few examples (shots) and suggest using the same shots for fair comparisons. State-of-the-art methods for few-shot cross-lingual learning follow the source-training + target-adapting paradigm. 
In this paper, we investigate deficiencies of this approach and propose more effective methods which significantly improve the transfer performance compared to target-adapting.

\subsection{Gradient Surgery}
Previous works on gradient optimization \citep{ZhaoChen2018GradNormGN,sener2018multi,yu2020gradient} have successfully utilized gradient-level techniques to improve the performance of multi-task models. In fact, mixed training multilingual data can be categorized into multi-task learning \citep{YuZhang2018AnOO} but in a monolithic manner by using a single language-agnostic objective on the concatenated data from all languages. Recently, multilingual machine translation utilizes gradient-level regularization to improve the translation performance \citep{wang-etal-2020-balancing,yang-etal-2021-improving-multilingual,ZiruiWang2021GradientVI}. In this paper, our experiments mainly focus on training multiple target languages, so we propose \textbf{stochastic gradient surgery} (Section \ref{sec:sgs}) which improves upon the original gradient surgery method \citep{yu2020gradient} to improve the overall performance.

\section{Methods}
\subsection{Ordinary Few-Shot Learning}
The current state-of-the-art few-shot cross-lingual transfer learning method \citep{lauscher-etal-2020-zero,hedderich-etal-2020-transfer,zhao-etal-2021-closer} includes two stages, \textit{source-training} and \textit{target-adapting}. In the source-training stage, a pre-trained LM such as mBERT \citep{devlin-etal-2019-bert} or XLM-R \citep{conneau-etal-2020-unsupervised} is fine-tuned with sufficient labeled data in the source language (which is usually English). In the target-adapting stage, the source-trained model is then fine-tuned only with a few examples in the target language. We abbreviate the name of this method to \textbf{ord-FS}.
\subsection{Mixed Fine-Tuning on All Target Languages}
\label{sec:all_target_lang}
The ord-FS method brings up a question: \textbf{is it necessary to fine-tune a language-specific model for each target language?} Can we use one model to handle all target languages to avoid excessively language-specific models? One straightforward method to have such a model is fine-tuning the source-trained model on concatenated examples of all target languages, instead of only one target. Here, we are interested in whether more few examples of other target languages will improve/degrade the overall performance. We abbreviate the name of this method to \textbf{mix-FT}.

\subsection{Mixed Training on Source and Target Languages}
\label{sec:naive-co-train}
Ord-FS and mix-FT follow the \textit{transductive transfer}\footnote{The pre-training (source-training) and the fine-tuning (target-adapting) are the same task.} learning method that first trains on the source domain and then fine-tunes on the target domain \citep{pan2009survey}. However, recently, \citet{xu-etal-2021-gradual} show that abruptly shifting the source domain to the target domain is not an optimized solution due to catastrophic forgetting \citep{mccloskey1989catastrophic}. Thus, we should be careful about the language domain gaps between the source and target languages, especially for distant languages. One naive but effective approach to preserve the source knowledge and escape catastrophic forgetting is simply training both the source and target data\footnote{Target data is randomly interpolated in the source data.} (all target languages), where we simplify source-training and target-adapting into only one mixed training step. We abbreviate the name of this method to \textbf{naive-mix-train}.

\subsection{Gradient Surgery in Mixed Training}
\label{sec:gradient-co-train}
One issue of naive-mix-train is conflicting gradients \citep{yu2020gradient} among languages, which makes training more difficult because gradients point away from one another. We define that two gradients are conflicting if they have a negative cosine similarity. Another issue is that the information of the target domain will be diluted due to the overwhelming source data. Specifically, the gradient of source data is much larger in magnitude than the other languages in one batch training due to the small or even no target training instances in this batch. Hence, the source gradients will dominate the average gradient and result in information dilution of the target data and underestimation of the target language performance.  

The main idea of using gradient surgery \citep{yu2020gradient} to mitigate the two issues above is, in each backpropagation step, projecting the dominant gradient to the normal plane of a target gradient to de-conflict their gradients and `\textit{remind}' the model of target instances. Specifically, we denote $g_s$ as the gradient for the source language and $g_t$ as the gradient for the target language. We first compute the cosine similarity between $g_s$ and $g_t$ and judge $g_s$ and $g_t$ are conflicting gradients if their similarity is negative. Next, we project $g_s$ into the normal plane of $g_t$ only if they are conflicting:

\begin{equation}
    g_s' = g_s-\frac{g_s\cdot g_t}{\parallel g_t\parallel^2}g_t
    \label{eq:gradient_surgery}
\end{equation}
The modified $g_s'$ replace the original dominant source gradient to update the model parameters.

\subsection{Stochastic Gradient Surgery}
\label{sec:sgs}
However, target data is usually not guaranteed to exist in the batch due to the small training size. Even though we assume that we have target data for all target languages in each batch training, we should detect conflicting gradients not just between source and target languages, but also between every target language. However, this is extremely computationally expensive, especially when it comes to large-scale languages for training. Based on this, we propose \textbf{stochastic gradient surgery} approach, composed of two parts, \textbf{oracle dataset creation} and \textbf{stochastic training}.

\paragraph{Oracle Dataset Creation} 
In the case of $K$-shot learning, the oracle dataset comprises $K$ training instances\footnote{XNLI use $K$ examples from every class followed by the ``N-way K-shot" discussion in Section \ref{sec:settings}.} for each target language. To not use any external information, the oracle datasets of target languages are exactly the $K$ target instances used in mixed training. Similar to \citet{wang-etal-2020-balancing,XinyiWang2021GradientguidedLM,yang-etal-2021-improving-multilingual}, we create an oracle dataset to ensure that we can pair any one of the target languages with the source language to operate gradient surgery. 

\paragraph{Stochastic Training}
In each batch training, we randomly pick oracle data of a random target language in a uniform distribution to conduct gradient surgery with the source batch data. Moreover, in order to avoid that small number of target examples constrain the source gradients into a sub-optimal place (especially for tasks which need higher-level semantic understanding), we also have a pre-set threshold $\alpha$ to control the probability of gradient surgery in each training step. The gradient surgery is conducted only if a sampled value $p\sim\text{uniform}[0,1]$ is smaller than $\alpha$. 

The advantages of this method are that 1) we only focus on gradient de-conflicting between the source and one of the target languages, which only computes the gradient one additional time to avoid expensive computation, 2) and more importantly, the source language could be a pivot language which also helps gradients of target languages de-conflict between each other (more discussion in Section \ref{sec:visual-deconflict} ). The detailed workflow is shown in Algorithm \ref{alg:sgs}. We abbreviate the name of this method to \textbf{gradient-mix-train}.


\begin{algorithm}[t!]
    \SetAlgoLined
    \DontPrintSemicolon
    \SetKwInOut{Input}{Input}
    \SetCommentSty{itshape}
    \SetKwComment{Comment}{$\triangleright$ }{}
    \Input{Language Set $\mathcal{L}$; Pre-Trained Model $\theta$; Mixed Training Data $\mathcal{D}_{\text{train}}$; Oracle Data $\mathcal{D}_{\text{oracle}}^l, l\in\mathcal{L}$; Pre-Set Threshold $\alpha$.}
    Initialize $\theta_0 = \theta$, step $t=0$\\
    \While{not converged}{
        \Comment{Iterate batches $\mathcal{B_{\text{train}}}$ from data $\mathcal{D}_{\text{train}}$}
        \For{$\mathcal{B_{\text{train}}}$ in $\mathcal{D}_{\text{train}}$}{
                $g_{\text{train}} = \nabla_{\theta_t} L(\theta_t, \mathcal{B_{\text{train}}})$ \\
                Sample a language $l$ from set $\mathcal{L}$\\
                $g_{\text{oracle}} = \nabla_{\theta_t} L(\theta_t, \mathcal{D}_{\text{oracle}}^l)$ \\
                Sample a value $p\sim\text{uniform}[0,1]$ \\
                \Comment{Gradient surgery}
                \If{$g_{\text{oracle}} \cdot g_{\text{train}} < 0$ and $p < \alpha$}{
                ${g}_{\text{train}} = g_{\text{train}} - \frac{g_{\text{train}} \cdot g_{\text{oracle}}}{\| g_{\text{oracle}} \|^2} ~ g_{\text{oracle}}$
            }
        Update $t \leftarrow t + 1$ \\
        Update $\theta_t$ with gradient ${g}_{\text{train}}$ \\
            }

    }
    \caption{Stochastic Gradient Surgery}
    \label{alg:sgs}
\end{algorithm}

\section{Experiments}
\label{sec:experiments}
\subsection{Development-Free Training}
\label{sec:dev_free_target}
Importantly, \citet{zhao-etal-2021-closer} notice that few-shot learning easily tends to overfit quickly at a small number of shots, where the model performs best on the dev set at the beginning of training. One good solution to avoid overfitting is using target dev set for early stopping. Previous studies \citep{hedderich-etal-2020-transfer,zhao-etal-2021-closer} utilize a large amount of dev sets for each target language for model selection, e.g., even around 10K dev examples for Arabic in the NER task. However, it is unlikely that such a dev set would be available in reality, especially for the extreme low-resource training such as 1-shot and 5-shot learning, since it would be more effective to use it for training instead \citep{kann-etal-2019-towards}.  The true standard setup of zero-shot cross-lingual learning only uses the source dev set \citep{zhao-etal-2021-closer}, and few-shot learning should also follow this setup, particularly at a small value of shots. Thus, we suggest \textbf{only using the source dev set for model selection}. However, for target-adapting, it does not makes sense to use the source dev for model selection due to the different languages in the training and dev steps. Hence, the two-step methods, ord-FS and mix-FT, use the last checkpoint for evaluation. Since naive-mix-train and gradient-mix-train train on both source and target data, they are suitable for using the source dev set for target model selection. We show that \textbf{our methods substantially outperform target-adapting whatever it uses unrealistic dev sets or not} in Section \ref{sec:results}.

We consider all introduced methods in the experiment, including two-step methods (ord-FS, mix-FT), and one-step methods (naive-mix-train, gradient-mix-train). Moreover, in order to investigate the difference between using and not using dev sets, we add another baseline, \textbf{ord-FS+dev}, whis is ord-FS with unrealistically large dev sets\footnote{Detail information of dev sets are shown in Appendix \ref{app:sec:dev_size}} for model selection as \citet{zhao-etal-2021-closer} conduct.
\begin{table*}[]
\centering
\resizebox{0.75\linewidth}{!}{
\begin{tabular}{l|l|cc|cc|cc|cc}
\hline
\multicolumn{1}{c|}{$K$} & \multicolumn{1}{c|}{Methods} & \multicolumn{2}{c|}{NER}                                                               & \multicolumn{2}{c|}{POS}                                                                                     & \multicolumn{2}{c|}{TyDiQA}                                                            & \multicolumn{2}{c}{XNLI}                                                           \\
                         &                              & Avg. F1 (\%)                                              & sd.                        & Avg. F1 (\%)                                              & sd.                                              & Avg. F1 (\%)                                              & sd.                        & Avg. Acc. (\%)                                        & sd.                        \\ \hline
$K=0$                    & Zero-Shot                    & 64.56                                                     & -                          & 77.32                                                     & -                                                & 55.80                                                     & -                          & 73.55                                                 & -                          \\ \hline
\multirow{5}{*}{$K=1$}   & ord-FS+dev \citep{zhao-etal-2021-closer}                   & \better 65.92                              & \small 0.84 & \muchbetter 80.37                          & \small 0.16                       & \better 55.81                              & \small 1.01 & \better 73.95                          & \small 0.19 \\
                         & ord-FS \citep{zhao-etal-2021-closer}                      & \worse 64.11                               & \small 0.98 & \better 80.24                              & \small 0.19                       & \muchworse 47.44                           & \small 1.47 & \better 73.70                          & \small 0.17 \\
                         & mix-FT (Ours)                 & \better 65.71                              & \small 0.90 & \better 79.37                              & \small 0.12                       & \muchworse 48.73                           & \small 2.15 & \worse 73.54                           & \small 0.61 \\
                         & naive-mix-train (Ours)        & \better 67.31                              & \small 0.58 & \better 80.04                              & \small 0.23                       & \better 57.03                              & \small 0.56 & \worse 73.29                           & \small 0.43 \\
                         & gradient-mix-train (Ours)     & \muchbetter\textbf{69.58} & \small 0.99 & \muchbetter\textbf{81.14} & \small 0.27                       & \better\textbf{57.64}     & \small 1.02 & \better\textbf{74.09} & \small 0.54 \\ \hline
\multirow{5}{*}{$K=5$}   & ord-FS+dev \citep{zhao-etal-2021-closer}                   & \muchbetter 68.22                          & \small 0.69 & \muchbetter 83.15                          & \small 0.23                       & \worse 55.60                               & \small 1.07 & \better 74.08                          & \small 0.36 \\
                         & ord-FS \citep{zhao-etal-2021-closer}                      & \better 65.91                              & \small 0.91 & \muchbetter 82.95                          & \small 0.20                       & \muchworse 51.19                           & \small 1.29 & \better 73.73                          & \small 0.60 \\
                         & mix-FT (Ours)                 & \muchbetter 70.60                          & \small 0.85 & \muchbetter 81.95                          & \small \small 0.16 & \worse 54.49                               & \small 1.76 & \worse 73.13                           & \small 0.74 \\
                         & naive-mix-train (Ours)        & \muchbetter 72.06                          & \small 0.68 & \muchbetter 82.79                          & \small \small 0.19 & \better 58.59                              & \small 1.45 & \better 73.69                          & \small 0.80 \\
                         & gradient-mix-train (Ours)     & \muchbetter\textbf{73.27} & \small 0.60 & \muchbetter\textbf{83.48} & \small \small 0.24 & \muchbetter\textbf{59.34} & \small 1.04 & \better\textbf{74.41} & \small 0.26 \\ \hline
\multirow{5}{*}{$K=10$}  & ord-FS+dev \citep{zhao-etal-2021-closer}                   & \muchbetter 69.85                          & \small 0.60 & \muchbetter 84.92                          & \small 0.07                       & \worse 55.59                               & \small 1.62 & \better 74.19                          & \small 0.39 \\
                         & ord-FS \citep{zhao-etal-2021-closer}                       & \muchbetter 68.75                          & \small 0.67 & \muchbetter 84.66                          & \small 0.08                       & \worse 53.17                               & \small 1.56 & \better 74.03                          & \small 0.38 \\
                         & mix-FT (Ours)                 & \muchbetter 73.89                          & \small 0.56 & \muchbetter 83.54                          & \small 0.07                       & \worse 55.54                               & \small 1.05 & \better 73.62                          & \small 0.98 \\
                         & naive-mix-train (Ours)        & \muchbetter 74.13                          & \small 0.45 & \muchbetter 84.52                          & \small 0.17                       & \muchbetter 58.88                          & \small 1.37 & \better 74.23                          & \small 0.37 \\
                         & gradient-mix-train (Ours)     & \muchbetter\textbf{75.92} & \small 0.61 & \muchbetter\textbf{85.03} & \small 0.16                       & \muchbetter\textbf{59.47} & \small 1.73 & \better\textbf{74.44} & \small 0.38 \\ \hline 
\end{tabular}
}
\caption{Main results of all methods with their standard deviation (sd.) of 5 repetitive experiments for all tasks with $K\in{1,5,10}$. Scores are averaged by all target languages. Best scores are \textbf{bold}. Cells are colored by performance difference over zero-shot baseline: \colorbox{muchbetter}{+3 or more}, \colorbox{better}{+0 to +3}, \colorbox{worse}{-0 to -3}, \colorbox{muchworse}{-3 or more}. \textbf{ord-FS+dev}: ordinary few-shot learning that fine-tunes on one target language each time with development set; \textbf{ord-FS}: the \textbf{ord-FS+dev} method without development set; \textbf{mix-FT}: mixed fine-tuning on concatenated target examples together; \textbf{naive-mix-train}: naively training both source and all target examples together; \textbf{gradient-mix-train}: utilizing stochastic gradient surgery during the naive-mix-train.}
\label{tab:main_results}
\end{table*}

\subsection{Tasks and Datasets}
 We consider two lower-level (structured prediction) tasks, Wikiann Named-Entity Recognition (NER) task \citep{ner} and Part-of-Speech Tagging (POS) \citep{nivre2018universal} and two different types of higher-level tasks, Typologically Diverse Question Answering-Gold Passage\footnote{We try to not use translated data such as XQuAD \citep{artetxe2020cross} to avoid unrealistic artifacts such as preserving source words \citep{Clark2020tydiqa}.} (TyDiQA-GoldP) \citep{Clark2020tydiqa} and Cross-lingual Natural Language Inference (XNLI) \citep{Conneau2018xnli}. We download datasets from the XTREME-R benchmark \citep{hu2020xtreme, ruder-etal-2021-xtreme}. NER and POS cover 48 and 38 languages, respectively. Our experiments use 35 languages on POS because the remaining three languages, Thai(\textit{th}), Tagalog(\textit{tl}) and Yoruba(\textit{yo}), do not have target training data in XTREME-R. TydiQA and XNLI cover 9 and  15 languages, respectively. We conduct aforementioned methods on all tasks for all languages. English is the source language and the others are targets. Statistics about languages are listed in Appendix \ref{app:lang}.

\subsection{Settings}
\label{sec:settings}
Two-step training methods, ord-FS(+dev) and mix-FT, have two different settings for source-training and target-adapting. For one-step methods, naive-mix-train and gradient-mix-train, their settings are the same as source-training in the two-step methods. We run 10 epochs for NER and POS, 60 for TyDiQA, and 10 for XNLI in both source-training and target-adapting. The batch size of all tasks is 32 for source-training and $K$ for target-adapting with a 2e-5 learning rate. Pre-set threshold $\alpha$ is 1 for NER and POS and 0.1 for TyDiQA and XNLI unless otherwise noted. The values of $\alpha$ are empirically selected, which might not be optimal but strongly effective. The model architecture of NER and POS is based on pre-trained XLM-R$_\texttt{large}$ attached with a feed-forward token-level classifier. For TydiQA, the representations of all subwords in XLM-R$_\texttt{base}$ are input to a span classification head –-- a linear layer computing the start and the end of the answer. For XNLI, the model architecture is XLM-R$_\texttt{base}$ with a simple softmax classifier on the vector of the start token. The number of examples we consider is $K\in\{1,5,10\}$. The sampling method is simply extracting random $K$ shots. The only exception is XNLI, where we adopt the sampling method of conventional few-shot classification learning --- ``$N$-way $K$-shot" \citep{fei2006one} --- we sample $K$ examples for $N$ classes. Here, $N$ is the total number of classes in XNLI. We repeat every experiment 5 times with 5 different random seeds\footnote{Shots are different with different seeds.} suggested by \citet{lauscher-etal-2020-zero}. All methods use the same $K$ shots for a fair comparison. We finally report the average accuracy (XNLI) or F1 scores (other tasks) and their standard deviation.

\subsection{Results}
\label{sec:results}
The main results on each task, conditioned on the number of examples $K$ and \textbf{averaged across all languages}, are presented in Table \ref{tab:main_results}. The full results with each target language are shown in Appendix \ref{app:full-results}. For all values of $K$ and all tasks, \textbf{gradient-mix-train performs the best among all introduced few-shot learning methods}.

The zero-shot cross-lingual transfer results ($K=0$) deliver similar results comparable to \citet{ruder-etal-2021-xtreme}. Similar to the findings in \citet{lauscher-etal-2020-zero,zhao-etal-2021-closer}, we notice substantial improvements with ord-FS(+dev) on lower-level tasks (NER and POS) and modest improvement on XNLI over zero-shot performance.

However, ord-FS significantly degrades the zero-shot performance on TyDiQA because it suffers from a tendency of overfitting on target training instances (more discussion in Section \ref{sec:dev-free}). On the other hand, with the help of dev sets in model selection, ord-FS+dev achieves higher performance than ord-FS on all tasks and particularly solve the overfitting issue.

Compared to ord-FS, NER and TyDiQA benefit most from mix-FT, e.g., from 65.91\% to 70.60\% with $K=5$ in NER. However, it still suffers from the overfitting issue, but the impact decrease with more target examples. Training source sentences with target data (naive-mix-train) is a better solution. It consistently outperforms mix-FT on all tasks with various $K$, and importantly, overcomes the serious overfitting on the TyDiQA task and highly boosts the performance (e.g., from 48.73\% of mix-FT to 57.03\% of naive-mix-train in 1-shot learning). Furthermore, applying stochastic gradient surgery on mixed training (gradient-mix-train) achieves the best performance on all tasks with all settings of $K$ and outperforms ord-FS by a significant margin, such as up to 7.36\% averaged absolute improvement on NER in 5-shot learning. On the other hand, the gap between our methods and ord-FS in POS is smaller than in NER (the same type of task). The reason could be that the strong POS task baseline has already left less room for further improvement. 

\begin{figure*}[ht]
     \centering
     
     \begin{subfigure}[b]{0.75\textwidth}
         \centering
         \resizebox{1\linewidth}{!}{
         \includegraphics[width=\textwidth]{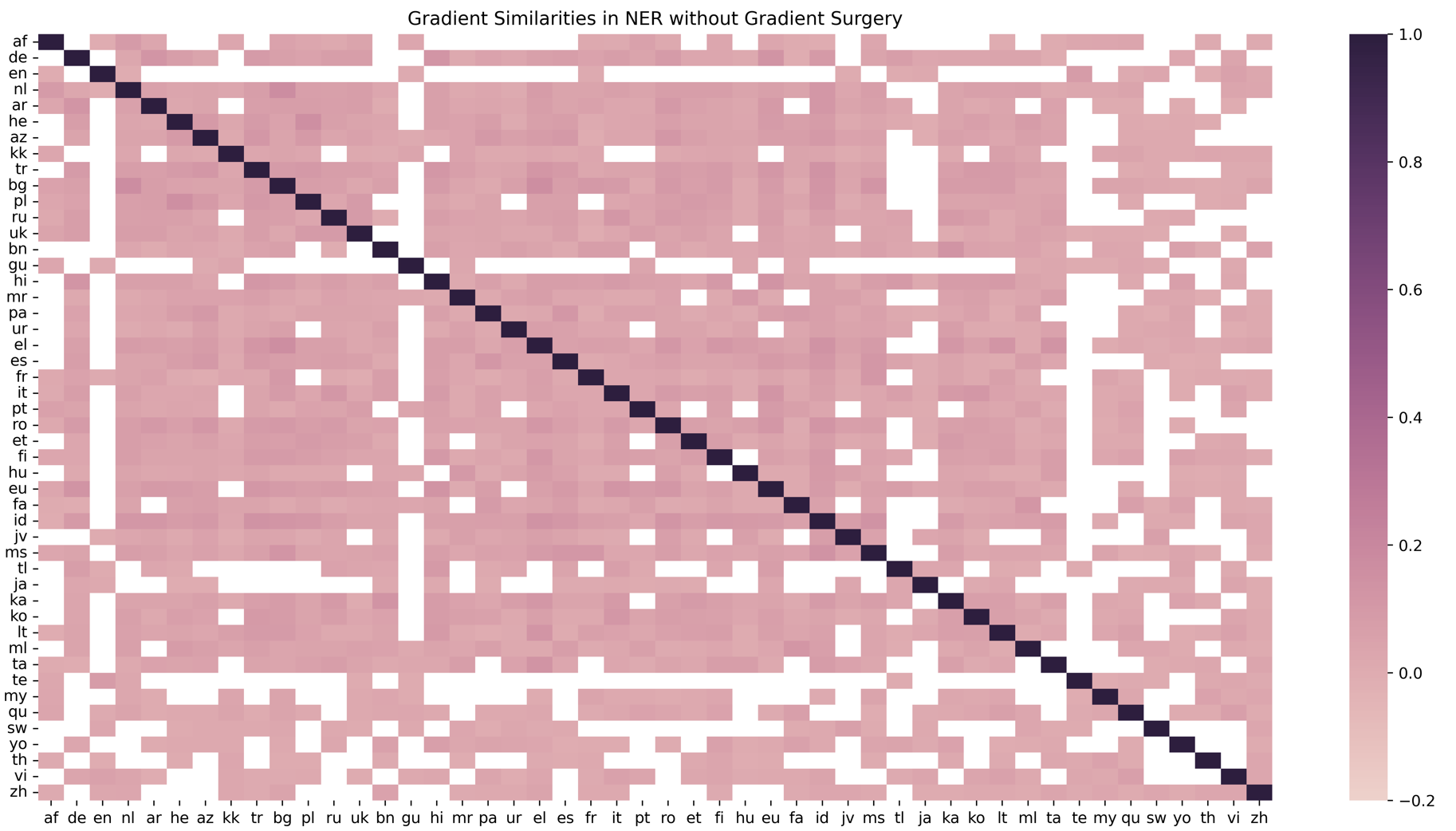}}
         \caption{Gradient similarity across languages without gradient surgery}
         \label{fig:no-grad}
     \end{subfigure}
     \hfill
     \begin{subfigure}[b]{0.75\textwidth}
         \centering
         \resizebox{1\linewidth}{!}{
         \includegraphics[width=\textwidth]{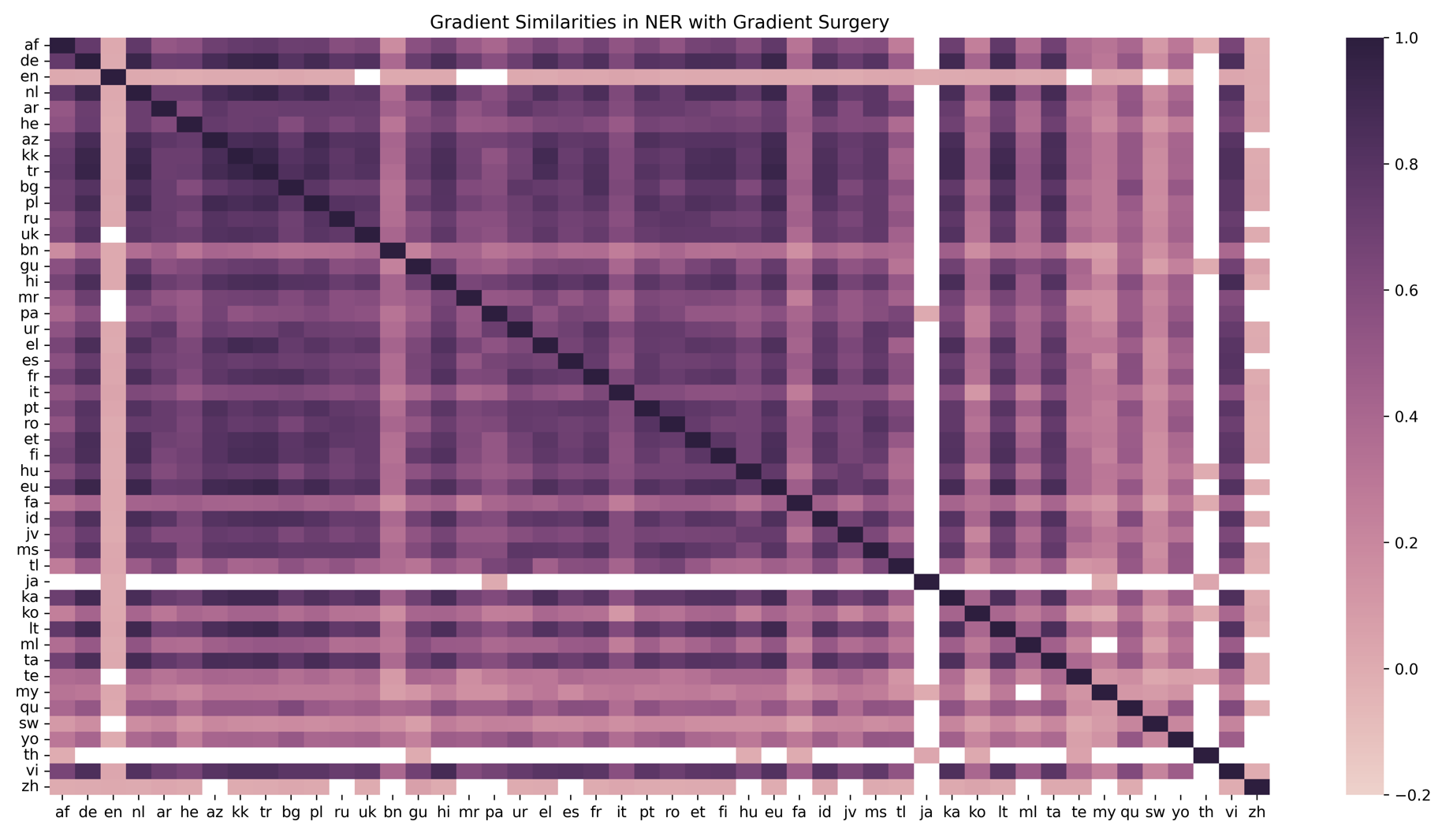}}
         \caption{Gradient similarity across languages with gradient surgery}
         \label{fig:with-grad}
     \end{subfigure}
     \caption{Gradient similarities across 48 languages in the NER task with 5 shots. Deeper colors represent higher cosine similarities. Conflicting gradients are directly marked as while cells in the heatmap. The similarities are highly improved after stochastic gradient surgery. The gradients are averaged from 5 checkpoints.}
     \label{fig:grad}
\end{figure*}

\begin{table}[]
\centering
\resizebox{1\linewidth}{!}{
\begin{tabular}{cc|cc|cc|cc}
\hline
\multicolumn{2}{c|}{NER} & \multicolumn{2}{c|}{POS} & \multicolumn{2}{c|}{TyDiQA} & \multicolumn{2}{c}{XNLI} \\
lang. & $\Delta$ F1 (\%) & lang.   & $\Delta$ F1 (\%)  & lang.   & $\Delta$ F1 (\%)  & lang. & $\Delta$ Acc. (\%) \\ \hline
pa    & 17.60            & wo      & 3.82              & bn      & 12.27             & sw    & 2.36             \\
zh    & 15.24            & mr      & 3.51              & te      & 11.14             & ur    & 1.95             \\
ar    & 14.14            & hi      & 2.60              & sw      & 10.58             & ru    & 1.68             \\
vi    & 13.22            & tr      & 2.18              & ar      & 9.45              & fr    & 0.91             \\
hi    & 12.68            & fi      & 1.55              & fi      & 9.05              & zh    & 0.78             \\ \hline
\end{tabular}}
\caption{Top-5 languages that achieve the highest improvement by using gradient-mix-train methods compared to ord-FS on all tasks in 5-shot learning. Most languages are distant from English. }
\label{tab:benefit_most_cotrain}
\end{table}

\section{Analysis and Discussion}
\subsection{Which Language Benefits Most?}
\label{sec:benefit_most}
Table \ref{tab:main_results} shows the strong effectiveness of gradient-mix-train in improving the overall performance of each task. Here, we are interested in taking a closer look at the results of specific languages and investigating which language benefits most. Take 5-shot learning as an example. Table \ref{tab:benefit_most_cotrain} illustrates the top-5 languages which boost most by using gradient-mix-train over ord-FS in all tasks\footnote{For the languages that benefit the least, gradient-mix-train still yields large gains over the baseline on NER and TyDiQA. We discuss this further in Appendix \ref{app:sec:benefits_least}.}, where the improvement is up to 17.60\% absolute F1 scores for \textit{pa} in the NER task. Most of the languages in the top-5 list are linguistically distant from English. We hypothesize that for such distant languages, the model has difficulty in learning the target training instances by abruptly shifting to the target domain. For closely related languages, the model is able to extrapolate the target-specific knowledge whose priors are close to English so that the model is less sensitive to these few target training examples than distant languages. However, gradient-mix-train is able to smoothly learn the distribution of source domain and extrapolate (distant) target domains by mixed training and gradient-level optimization.

\begin{figure*}[ht]
     \centering
     \begin{subfigure}[b]{0.6\textwidth}
         \centering
         \resizebox{1\linewidth}{!}{
         \includegraphics[width=\textwidth]{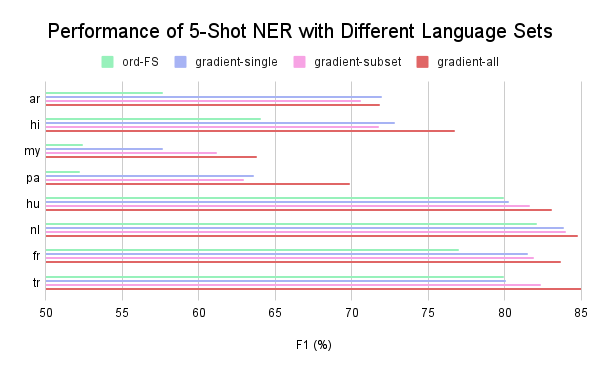}}
         \caption{Performance on various subsets of languages in NER}
         \label{fig:ner-subset}
     \end{subfigure}
     \hfill
     \begin{subfigure}[b]{0.6\textwidth}
         \centering
         \resizebox{1\linewidth}{!}{
         \includegraphics[width=\textwidth]{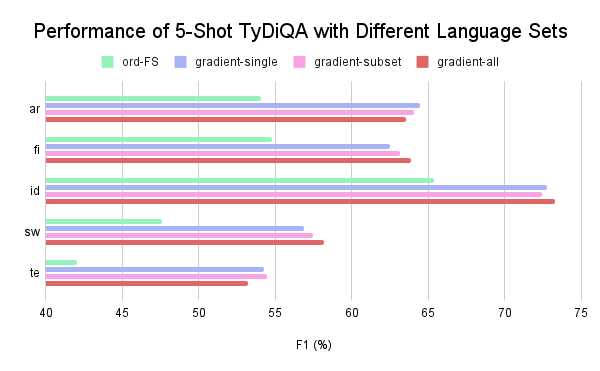}}
         \caption{Performance on various subsets of languages in TyDiQA}
         \label{fig:tydiqa-subset}
     \end{subfigure}
     \caption{Performance of gradient-mix-train on different sets of languages compared to ord-FS for \textbf{(a)} NER and \textbf{(b)} TyDiQA. \textbf{gradient-\{all,subset,single\}} represents training on all/subset/single languages by using graident-mix-train.}
     \label{fig:subset}
\end{figure*}

\subsection{Visualization of Gradient De-Conflicting}
\label{sec:visual-deconflict}
We take the NER task as an example to analyze the gradient de-conflicting of stochastic gradient surgery since it covers the most languages among all tasks. In Figure \ref{fig:grad}, we use a symmetric heatmap to visualize pair-wise gradient similarities, averaged by all 5 checkpoints in 5-shot learning. Note that languages in the figure are adjacent to other languages in the same linguistic language family. The gradient of English is calculated by the randomly picked 100 batches on average, and gradients of the other target languages are calculated by their 5 training instances. To highlight the conflicting gradients across languages, we directly mark the cells with negative similarities as pure white color. 

Figure \ref{fig:no-grad} shows the gradient similarities of the naive-mix-train model. As expected, gradient similarities of many language pairs are conflicting (white color cells), and gradients of most languages are approximately orthogonal, where their similarities are close to 0. It is worth mentioning that gradients similarities between English and most languages are conflicting. In comparison, in Figure \ref{fig:with-grad}, we illustrates the gradient similarities of gradient-mix-train, and the gradient similarities between English and most of the target languages are positive. Moreover, gradients of most target language pairs have higher similarities (deeper colors), which also verifies the correctness of our statement that target languages utilize English as a pivot language to de-conflict and even improve their similarities. The only two exceptions are \textit{th} and \textit{ja}, the two hardest task in NER, whose F1 in zero-shot learning is only 1.02\% and 18.31\%. Their similarities with other languages are negative but positive between themselves. However,  gradient-mix-train still achieve impressive improvement on \textit{th} ($\Delta=3.13\%$) and \textit{ja} (\textit{$\Delta=5.40\%$}) compared to naive-mix-train (see the full results in Appendix \ref{app:full-results}). 
We also notice the clustering by membership closeness in the linguistic family, along with the diagonal of gradient similarity matrix, e.g., Indo-Aryan (\textit{bn}, \textit{gu}, \textit{hi}, \textit{mr}, \textit{pa}, \textit{ur}). Moreover, some language families are positive correlated, e.g., Slavic(\textit{bg}, \textit{pl}, \textit{ru}, \textit{uk}) and Austronesian(\textit{id}, \textit{jv}, \textit{ms}).

\begin{figure}[ht]
    \centering
    \resizebox{1\linewidth}{!}{
    \includegraphics[width=7.5cm]{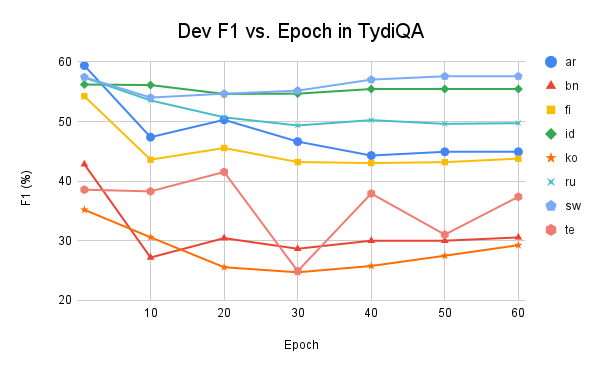}}
    \caption{Dev F1 scores of ord-FS+dev in TyDiQA. 6 out of 8 target languages overfit quickly, where they achieve the best performance at the first epoch.}
    \label{fig:dev-epoch}
\end{figure}

\subsection{Mixed Training with One Single Language}
\label{sec:cotrain-single}
In some cases, people are only interested in one target language and do not have resources for other languages. Hence, we further explore the effectiveness of gradient-mix-train in one target language case. We conduct experiments on the NER and TyDiQA tasks that show larger gaps among different methods than other two tasks. Considering the high expense of training the source data from scratch for every target language, we run experiments on subsets of languages for each task. For the NER task, we test on 8 languages: \textit{ar}, \textit{hi}, \textit{my}, \textit{pa}, which use different scripts from English, and \textit{hu}, \textit{nl}, \textit{fr}, \textit{tr}, which share the same script with English. Figure \ref{fig:ner-subset} shows the results for NER. Gradient-mix-train with only one single language is labeled as \textbf{gradient-single}\footnote{We reduce $\alpha$ for NER to 0.1 due to only one language considered.} in the figure (blue, the second bar). We can focus on comparing ord-FS (green, the first bar). We notice that gradient-single still outperforms ord-FS by a large margin for 4 non-Latin-script languages (e.g., 14.29\% improvement for \textit{ar}). In comparison, their gap becomes smaller when it comes to 4 Latin-script languages (e.g., 1.78\% improvement for \textit{nl}). Numeric results are shown in Appendix \ref{app:co-train-subset}. For the TyDiQA task, We pick 5 languages: \textit{ar}, \textit{fi}, \textit{id}, \textit{sw}, \textit{te}. We note that gradient-single still highly boosts the performance compared to ord-FS.

\subsection{Do the Same Scripts Help?}
\label{sec:co-train-subset}
Continuing the previous discussions in Section \ref{sec:cotrain-single}, we add a new baseline, \textbf{gradient-all} (red, the last bar in Figure \ref{fig:subset}), which uses gradient-mix-train method with all languages (original settings). Interestingly, gradient-all outperforms gradient-single on all selected languages except for \textit{ar} in NER, and a similar phenomenon also happens in TyDiQA. Note that \textit{ar} is the only language that uses Arabic script in TyDiQA and only shares the same script with \textit{yo} and \textit{kk} among 48 languages in NER. It brings a question that \textbf{do small examples of other languages which use the same scripts help in few-shot learning?} Hence, we move our experiments further on using gradient-mix-train with subsets of languages. We still consider the languages used in Section \ref{sec:cotrain-single}. Note that these languages are carefully selected. In NER, only \textit{my} and \textit{pa} share the same script (Brahmic) among 4 distant languages, and \textit{hu}, \textit{nl}, \textit{fr}, \textit{tr} share the Latin script from different language families. In TyDiQA, only \textit{fi},\textit{id} and \textit{sw} use the same script (Latin). We train 4 similar languages and 4 distant languages in NER, respectively. For TyDiQA, we train all 5 languages. The results of mixed training on subset of languages is denoted as \textbf{gradient-subset}\footnote{$\alpha$ is 0.4 for NER to ensure that each language has the same chance of explosion as gradient-single during training.} (pink, the third bar) in Figure \ref{fig:subset}. As expected, gradient-subset achieves better performance than gradient-single on all similar languages and on \textit{my} among distant languages in the NER task. As for other languages using distinct scripts, their performance slightly degenerates compared to gradient-single. A similar discussion also holds for the high-level TyDiQA task, but gaps between gradient-single and gradient-subset are smaller. In conclusion, to pursue the best performance, we recommend using gradient-mix-train with languages that share the same script or only a single language that uses a distinct script. 

\subsection{Escaping from Overfitting}
\label{sec:dev-free}
The overfitting causes the significant degeneration of ord-FS performance in TydiQA. Figure \ref{fig:dev-epoch} shows that 6 out of 8 target languages achieve the best dev score at the first epoch and decrease significantly afterwards. However, the phenomenon of degeneration is imperceptible in other tasks because only a few languages hit the same overfitting issue, e.g., 6.38\% of languages achieve the best score at the first epoch in 1-shot learning for NER, and none of them has the issue in 10-shot learning. Different from two-step methods, one of the biggest benefits of gradient-mix-train is the perfect fit for only using the source dev set to avoid overfitting (for model selection) because training and dev steps use the same (dominant) language. Thus, although gradient-mix-train can also be further improved by using unrealistic target dev sets, the gaps are smaller compared to ord-FS (Appendix \ref{app:sec:unrealistic_dev}).

%

\section{Conclusion}
We study the deficiencies of target-adapting in few-shot cross-lingual transfer and propose a mixed training method with gradient-level optimization. Our best model achieves state-of-the-art on four diverse NLP tasks with all values of $K$. Moreover, we are the first to use a single model to train all target languages and find that languages can benefit from others that share the same scripts. We also show the effectiveness of our method compared to target-adapting in a single target language case, and the gaps are still significant. Finally, we propose only using source dev set in few-shot settings and show that our method is development-free for targets and also able to escape from overfitting issues.

\section*{Acknowledgements}
We thank anonymous reviewers for their valuable comments. We thank Kelly Marchisio for her helpful suggestions. This work was supported in part by IARPA BETTER (\#2019-19051600005). The views and conclusions contained in this work are those of the authors and should not be interpreted as necessarily representing the official policies, either expressed or implied, or endorsements of ODNI, IARPA, or the U.S. Government. The U.S. Government is authorized to reproduce and distribute reprints for governmental purposes notwithstanding any copyright annotation therein.

\bibliography{anthology,custom}
\bibliographystyle{acl_natbib}

\clearpage
\appendix

\section{Size of Dev Sets}
\label{app:sec:dev_size}
\begin{figure*}[h]
    \centering
    \resizebox{1\linewidth}{!}{
    \includegraphics[width=8.5cm]{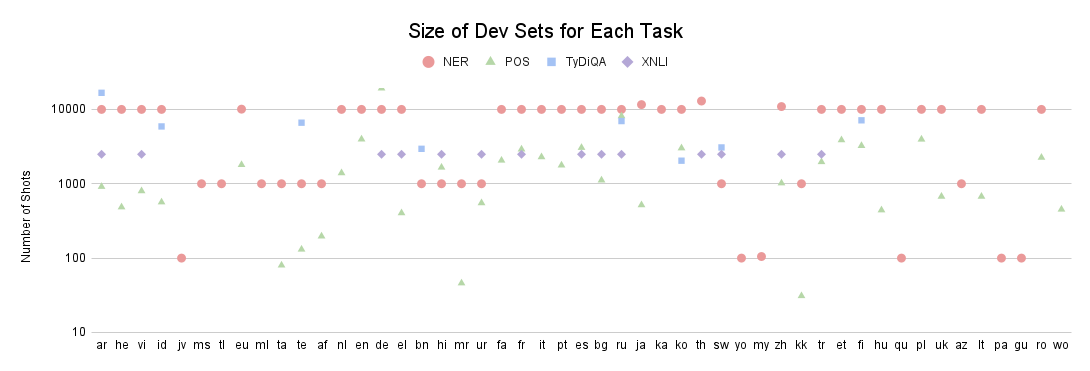}}
    \caption{The size of dev sets that we use in the experiments for each language in each task.}
    \label{app:fig:devsize}
\end{figure*}
In Figure \ref{app:fig:devsize}, we show the size of dev sets that we used in our experiments, which are also the dev sets used by \citet{zhao-etal-2021-closer}. Data of all tasks are downloaded from the XTREME-R benchmark \citep{hu2020xtreme,ruder-etal-2021-xtreme}, where train/dev/test sets are already split. We can notice that the dev size of all languages in all tasks are tremendously higher than the largest number (10) of shots we pick in few-shot cross-lingual learning. However, in reality, if we only have access to a few training instance, we usually do not have a such large dev set. For tasks such as NER, POS and XNLI, we sample shots from the target training sets and directly use their supported dev sets. For TyDiQA which only supports train and dev sets in XTREME-R, we sample shots from the target training sets but use the remaining training data as dev sets, and we use dev sets for test.

\section{Language Statistics}
\label{app:lang}
\begin{table*}[]
\centering
\resizebox{1\linewidth}{!}{
\begin{tabular}{lcllcccc}
\hline
Langugae     & ISO 639-1 code & Script       & Language Family & NER       & POS       & TyDiQA    & XNLI      \\ \hline
Afrikaans    & af        & Latin        & IE:Germanic     & \checkmark & \checkmark &           & \checkmark \\
Arabic       & ar        & Arabic       & Afro-Asiatic    & \checkmark & \checkmark & \checkmark &           \\
Azerbaijani  & az        & Latin        & Turkic          & \checkmark &           &           &           \\
Bulgarian    & bg        & Cyrillic     & IE:Slavic       & \checkmark & \checkmark &           & \checkmark \\
Bengali      & bn        & Brahmic      & IE:Indo-Aryan   & \checkmark &           & \checkmark &           \\
German       & de        & Latin        & IE:Germanic     & \checkmark & \checkmark &           & \checkmark \\
Greek        & el        & Greek        & IE:Greek        & \checkmark & \checkmark &           & \checkmark \\
English      & en        & Latin        & IE:Germanic     & \checkmark & \checkmark & \checkmark & \checkmark \\
Spanish      & es        & Latin        & IE:Romance      & \checkmark & \checkmark &           & \checkmark \\
Estonian     & et        & Latin        & Uralic          & \checkmark & \checkmark &           &           \\
Basque       & eu        & Latin        & Basque          & \checkmark & \checkmark &           &           \\
Persian      & fa        & Perso-Arabic & IE:Iranian      & \checkmark & \checkmark &           &           \\
Finnish      & fi        & Latin        & Uralic          & \checkmark & \checkmark & \checkmark &           \\
French       & fr        & Latin        & IE:Romance      & \checkmark & \checkmark &           & \checkmark \\
Gujarati     & gu        & Brahmic      & IE:Indo-Aryan   & \checkmark &           &           &           \\
Hebrew       & he        & Jewish       & Afro-Asiatic    & \checkmark & \checkmark &           &           \\
Hindi        & hi        & Devanagari   & IE:Indo-Aryan   & \checkmark & \checkmark &           & \checkmark \\
Hungarian    & hu        & Latin        & Uralic          & \checkmark & \checkmark &           &           \\
Indonesian   & id        & Latin        & Austronesian    & \checkmark & \checkmark & \checkmark &           \\
Italian      & it        & Latin        & IE:Romance      & \checkmark & \checkmark &           &           \\
Japanese     & ja        & Ideograms    & Japonic         & \checkmark & \checkmark &           &           \\
Javanese     & jv        & Brahmic      & Austronesian    & \checkmark &           &           &           \\
Georgian     & ka        & Georgian     & Kartvelian      & \checkmark &           &           &           \\
Kazakh       & kk        & Arabic       & Turkic          & \checkmark & \checkmark &           &           \\
Korean       & ko        & Hangul       & Koreanic        & \checkmark & \checkmark & \checkmark &           \\
Lithuanian   & lt        & Latin        & IE:Baltic       & \checkmark & \checkmark &           &           \\
Malayalam    & ml        & Brahmic      & Dravidian       & \checkmark &           &           &           \\
Marathi      & mr        & Devanagari   & IE:Indo-Aryan   & \checkmark & \checkmark &           &           \\
Malay        & ms        & Latin        & Austronesian    & \checkmark &           &           &           \\
Burmese      & my        & Brahmic      & Sino-Tibetan    & \checkmark &           &           &           \\
Dutch        & nl        & Latin        & IE:Germanic     & \checkmark & \checkmark &           &           \\
Punjabi      & pa        & Brahmic      & IE:Indo-Aryan   & \checkmark &           &           &           \\
Polish       & pl        & Latin        & IE:Slavic       & \checkmark & \checkmark &           &           \\
Portuguese   & pt        & Latin        & IE:Romance      & \checkmark & \checkmark &           &           \\
CuscoQuechua & qu        & Latin        & Quechuan        & \checkmark &           &           &           \\
Romanian     & ro        & Latin        & IE:Romance      & \checkmark & \checkmark &           &           \\
Russian      & ru        & Cyrillic     & IE:Slavic       & \checkmark & \checkmark & \checkmark & \checkmark \\
Swahili      & sw        & Latin        & Niger-Congo     & \checkmark &           & \checkmark & \checkmark \\
Tamil        & ta        & Brahmic      & Dravidian       & \checkmark & \checkmark &           &           \\
Telugu       & te        & Brahmic      & Dravidian       & \checkmark & \checkmark & \checkmark &           \\
Thai         & th        & Brahmic      & Kra-Dai         & \checkmark &           &           & \checkmark \\
Tagalog      & tl        & Brahmic      & Austronesian    & \checkmark &           &           &           \\
Turkish      & tr        & Latin        & Turkic          & \checkmark & \checkmark &           & \checkmark \\
Ukrainian    & uk        & Cyrillic     & IE:Slavic       & \checkmark & \checkmark &           &           \\
Urdu         & ur        & Perso-Arabic & IE:Indo-Aryan   & \checkmark & \checkmark &           & \checkmark \\
Vietnamese   & vi        & Latin        & Austro-Asiatic  & \checkmark & \checkmark &           & \checkmark \\
Wolof        & wo        & Latin        & Niger-Congo     &           & \checkmark &           &           \\
Yoruba       & yo        & Arabic       & Niger-Congo     & \checkmark &           &           &           \\
Mandarin     & zh        & Chinese      & ideograms       & \checkmark & \checkmark &           & \checkmark \\ \hline
\end{tabular}
}
\caption{Statistics about languages considered in this paper, including the scripts and language family of every language. A language used in a task is checkmarked under the column of the task.}
\label{app:tab:full-lang}
\end{table*}
In this paper, we cover a total of 49 languages in our whole experiments, including NER, POS, TyDiQA, and XNLI tasks. The list of full names of languages is shown in Table \ref{app:tab:full-lang}, with their ISO 639-1 code, script, and language families. We checkmark under the column of the task in the Table if the language is involved in the task.

\section{Full Results}
\label{app:full-results}
\begin{table*}[]
\centering
\resizebox{1\linewidth}{!}{
\begin{tabular}{llccccccccccccccccccccccccc}
\hline
$K$                     & Methods           & ar                              & he                              & vi                              & id                              & jv                              & ms                              & tl                              & eu                              & ml                              & ta                              & te                              & af                              & nl                              & en                              & de                              & el                              & bn                              & hi                              & mr                              & ur                              & fa                              & fr                              & it                              & pt                              & es                              \\ \hline
$K=0$                   & Zero-Shot         & 45.75                           & 55.35                           & 78.67                           & 52.47                           & 61.35                           & 69.65                           & 71.95                           & 56.37                           & 65.79                           & 55.82                           & 52.85                           & 78.34                           & 83.76                           & 84.50                           & 78.78                           & 78.38                           & 74.39                           & 69.71                           & 61.87                           & 54.85                           & 56.82                           & 79.78                           & 81.39                           & 81.91                           & 76.64                           \\
\hline
\end{tabular}
}
\caption{Full results (F1) of the NER task.}
\label{app:tab:full_ner}
\end{table*}

\begin{table*}[]
\centering
\resizebox{1\linewidth}{!}{
\begin{tabular}{llcccccccccccccccccc}
\hline
$K$                     & Methods           & af                              & ar                              & bg                              & de                              & el                              & en                              & es                              & et                              & eu                              & fa                              & fi                              & fr                              & he                              & hi                              & hu                              & id                              & it                              & ja                              \\ \hline
$K=0$                   & Zero-Shot         & 89.39                           & 69.52                           & 88.65                           & 88.50                           & 86.40                           & 96.12                           & 89.18                           & 86.74                           & 73.20                           & 74.49                           & 86.22                           & 87.79                           & 68.91                           & 75.50                           & 83.75                           & 83.32                           & 89.63                           & 27.82                           \\
\hline
\end{tabular}
}
\caption{Full results (F1) of the POS task.}
\label{app:tab:full_udpos}
\end{table*}

\begin{table*}[]
\centering
\resizebox{1\linewidth}{!}{
\begin{tabular}{llcccccccccc}
\hline
$K$                     & Methods           & ar                              & bn                              & fi                              & id                              & ko                              & ru                              & sw                              & te                              & en                              & Avg.                            \\ \hline
$K=0$                   & Zero-Shot         & 62.53                           & 42.24                           & 61.82                           & 70.62                           & 42.99                           & 57.75                           & 56.40                           & 43.23                           & 65.51                           & 55.80                           \\ \hline
\multirow{5}{*}{$K=1$}  & ord-FS+dev         & 62.20                           & \textbf{45.92}                  & 59.33                           & 71.15                           & 38.70                           & 58.70                           & 53.63                           & 47.12                           & 65.51                           & 55.81                           \\
                        & ord-FS            & 48.53                           & 33.91                           & 54.56                           & 63.36                           & 40.43                           & 49.59                           & 47.56                           & 23.49                           & 65.51                           & 47.44                           \\
                        & mix-FT             & 50.93                           & 36.83                           & 54.04                           & 61.72                           & 38.46                           & 51.35                           & 46.26                           & 33.48                           & 65.51                           & 48.73                           \\
\hline
\end{tabular}
}
\caption{Full results (F1) of the TyDiQA task.}
\label{app:tab:full_tydiqa}
\end{table*}

\begin{table*}[]
\centering
\resizebox{1\linewidth}{!}{
\begin{tabular}{llcccccccccccccccc}
\hline
$K$                     & Methods           & ar                              & bg                              & de             & el                              & es                              & fr                              & hi                              & ru                              & sw                              & th                              & tr                              & ur                              & vi                              & zh                              & en                              & Avg.                            \\ \hline
$K=0$                   & Zero-Shot         & 72.28                           & 77.15                           & 75.97          & 74.71                           & 78.56                           & 77.19                           & 69.10                           & 73.95                           & 62.08                           & 71.52                           & 72.32                           & 65.39                           & 74.15                           & 73.67                           & 85.19                           & 73.55                           \\ \hline
\multirow{5}{*}{$K=1$}  & ord-FS+dev         & \textbf{72.08}                  & 77.49                           & \textbf{76.19} & \textbf{75.47}                  & \textbf{79.21}                  & \textbf{77.99}                  & 69.16                           & 74.69                           & 62.29                           & 72.31                           & \textbf{72.46}                  & 66.00                           & 74.60                           & 74.20                           & \textbf{85.19} & 73.95                           \\
\hline
\end{tabular}
}
\caption{Full results (accuracy) of the XNLI task.}
\label{app:tab:full_xnli}
\end{table*}
The full results of NER, POS, TyDiQA and XNLI are shown in Table \ref{app:tab:full_ner}, Table \ref{app:tab:full_udpos}, Table \ref{app:tab:full_tydiqa} and Table \ref{app:tab:full_xnli}, respectively. In each task, we report F1 scores (or accuracy) of all covered languages in 1-,5-, or 10 shot learning by using all introduced methods. Best score among methods in each language is \textbf{bold}.

\section{Languages Benefits Least}
\label{app:sec:benefits_least}
In Table \ref{app:tab:benefit_least_cotrain}, we show the list of top-5 language which benefits least by using gradient-mix-train in 5-shot learning. In NER and XNLI, we can notice a reverse phenomenon in the top-5 languages which benefit most ---  most of the languages are linguistically closer to English, at least using the same (Latin) script. In NER and TyDiQA tasks, although the gap left by gradient-mix-train is much smaller than top-5 languages which benefits most, the improvements are still significant.
\begin{table*}[]
\centering
\begin{tabular}{cc|cc|cc|cc}
\hline
\multicolumn{2}{c|}{NER} & \multicolumn{2}{c|}{POS} & \multicolumn{2}{c|}{TyDiQA} & \multicolumn{2}{c}{XNLI} \\
lang. & $\Delta$ F1 (\%) & lang.   & $\Delta$ F1 (\%)  & lang.   & $\Delta$ F1 (\%)  & lang. & $\Delta$ Acc. (\%) \\ \hline
pl    & 0.90             & vi      & -4.35             & ko      & 5.25              & es    & -0.32            \\
ms    & 1.51             & ja      & -1.85             & ru      & 5.62              & tr    & -0.31            \\
nl    & 2.66             & he      & -1.36             & id      & 7.90              & de    & 0.23             \\
af    & 2.73             & hu      & -1.02             & fi      & 9.05              & vi    & 0.34             \\
sw    & 2.76             & zh      & -0.71             & ar      & 9.45              & fr    & 0.36             \\ \hline
\end{tabular}
\caption{Top-5 languages that achieve the least improvement by using gradient-mix-train compared to ord-FS on all tasks in 5-shot learning.}
\label{app:tab:benefit_least_cotrain}
\end{table*}

\section{Mixed Training with Subsets of Languages}
\label{app:co-train-subset}
\begin{table*}[]
\centering
\begin{tabular}{l|cccccccc}
\hline
\multicolumn{1}{c|}{} & ar    & hi    & my    & pa    & hu    & nl    & fr    & tr    \\ \hline
ord-FS                & 57.69 & 64.08 & 52.44 & 52.28 & 79.91 & 82.11 & 76.99 & 79.97 \\
gradient-single       & 71.98 & 72.82 & 57.67 & 63.60  & 80.28 & 83.89 & 81.51 & 80.10  \\
gradient-subset       & 70.57 & 71.79 & 61.18 & 62.99 & 81.65 & 83.96 & 81.90  & 82.37 \\
gradient-all          & 71.83 & 76.76 & 63.78 & 69.88 & 83.06 & 84.77 & 83.64 & 85.21 \\ \hline
\end{tabular}
\caption{Numeric results of Figure \ref{fig:ner-subset}.}
\label{app:tab:co-train-ner}
\end{table*}

\begin{table*}[]
\centering
\begin{tabular}{l|ccccc}
\hline
\multicolumn{1}{c|}{} & ar    & fi    & id    & sw    & te    \\ \hline
ord-FS                & 54.07 & 54.82 & 65.39 & 47.61 & 42.04 \\
gradient-single       & 64.43 & 62.52 & 72.78 & 56.91 & 54.28 \\
gradient-subset       & 64.04 & 63.17 & 72.44 & 57.48 & 54.49 \\
gradient-all          & 63.52 & 63.87 & 73.29 & 58.20 & 53.19 \\ \hline
\end{tabular}
\caption{Numeric results of Figure \ref{fig:tydiqa-subset}.}
\label{app:tab:co-train-tydiqa}
\end{table*}
Here, we show the numeric results of Figure \ref{fig:ner-subset} and Figure \ref{fig:tydiqa-subset} in Table \ref{app:tab:co-train-ner} and Table \ref{app:tab:co-train-tydiqa}, respectively.

\section{Our Methods with Dev Sets}
\label{app:sec:unrealistic_dev}
\begin{figure*}[ht]
    \centering
    \includegraphics[width=12cm]{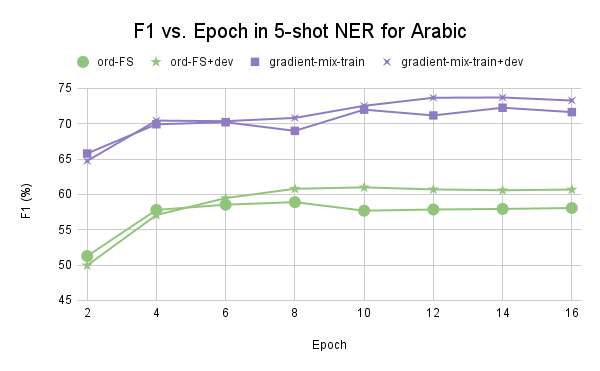}
    \caption{F1 scores of gradient-mix-train(+dev) and ord-FS(+dev) with increasing number of epochs. The large dev set helps model selection after certain epochs. Gradient-mix-train shows less gap led by the dev set than ord-FS and can select approximately optimal model by only using the source dev set.}
    \label{app:fig:arabic-ner}
\end{figure*}

We take \textit{ar} in the NER task as an example to show that gradient-mix-train can be further improved by utilizing large dev sets (around 10K). Figure \ref{app:fig:arabic-ner} shows F1 scores of gradient-mix-train and ord-FS both with and without dev sets with increasing epoch numbers. Methods with the help of the dev set start showing its effectiveness in model selection when it comes to large enough epoch numbers. Importantly, the gap led by the dev set in gradient-mix-train is smaller than the one in ord-FS, which shows that gradient-mix-train is able to select approximately optimal model even without target dev sets by using the source dev set. It is also worth mentioning that gradient-mix-train even significantly outperforms the best performance of ord-FS with only 2 epoch of source (and target) data training. Still, ord-FS starts training based on 10-epoch source-trained model. 

\end{document}